\documentclass{article}

\usepackage{arxiv}

\usepackage[utf8]{inputenc} 
\usepackage[T1]{fontenc}    
\usepackage{hyperref}       
\usepackage{url}            
\usepackage{booktabs}       
\usepackage{amsfonts}       
\usepackage{nicefrac}       
\usepackage{microtype}      
\usepackage{lipsum}
\usepackage{graphicx}
\usepackage{indentfirst}
\graphicspath{ {./images/} }
\usepackage[table,xcdraw]{xcolor}
\usepackage{enumerate}
\usepackage[shortlabels]{enumitem}

\usepackage{authblk}

\usepackage{wrapfig}
\usepackage{graphicx}
\usepackage{picinpar}

\makeatletter
\long\def\figwindownonum[#1,#2,#3,#4] {
  \begin{window}[#1,#2,{#3},{\centering#4\par}] }
\def\endfigwindownonum{\end{window}}
\makeatother

\title{The 3rd Anti-UAV Workshop \& Challenge: \\
Methods and Results}

\author[1,18,19]{Jian Zhao}
\author[2]{Jianan Li}
\author[3]{Lei Jin}
\author[3]{Jiaming Chu}
\author[1]{Jun Wang}
\author[1]{Jiangqiang Xia}
\author[4]{Kai Wang}
\author[5]{Yang Liu}
\author[6]{Sadaf Gulshad}
\author[7]{Jiaojiao Zhao}
\author[8]{Tianyang Xu}
\author[8]{Xuefeng Zhu}
\author[8]{Shihan Liu}
\author[9]{Zheng Zhu}
\author[10]{Guibo Zhu}
\author[11]{Zechao Li}
\author[12]{Zheng Wang}
\author[13]{Baigui Sun}
\author[14]{Yandong Guo}
\author[15]{Shin’ichi Satoh}
\author[16]{Junliang Xing}
\author[17]{Jane Shen Shengmei}

\affil[1]{National Innovation Institute of Defense Technology}
\affil[2]{Beijing Institute of Technology}
\affil[3]{Beijing University of Posts and Telecommunication}
\affil[4]{National University of Singapore}
\affil[5]{Alibaba Inc.}
\affil[6]{Bosch Delta Lab, University of Amsterdam}
\affil[7]{VIS Lab, University of Amsterdam}
\affil[8]{Jiangnan University}
\affil[9]{PhiGent Robotics}
\affil[10]{Institute of Automation, Chinese Academy of Sciences}
\affil[11]{Nanjing University of Science and Technology}
\affil[12]{Wuhan University}
\affil[13]{Alibaba DAMO Academy}
\affil[14]{IDEA Visiting Researcher}
\affil[15]{National lnstitute of lnformatics and University of Tokyo}
\affil[16]{Tsinghua University}
\affil[17]{Pensees Singapore}
\affil[18]{Intelligent Game and Decision Laboratory}
\affil[19]{Department of Mathematics and Theories, Peng Cheng Laboratory}

\begin{document}
\maketitle
\begin{abstract}
The 3rd Anti-UAV Workshop \& Challenge aims to encourage research in developing novel and accurate methods for multi-scale object tracking. The Anti-UAV dataset used for the Anti-UAV Challenge has been publicly released. There are two main differences between this year's competition and the previous two. First, we have expanded the existing dataset, and for the first time, released a training set so that participants can focus on improving their models. Second, we set up two tracks for the first time, \emph{i.e.}, Anti-UAV Tracking and Anti-UAV Detection \& Tracking. Around 76 participating teams from the globe competed in the 3rd Anti-UAV Challenge. In this paper, we provide a brief summary of the
3rd Anti-UAV Workshop \& Challenge including brief introductions to the top three methods in each track. The submission leaderboard will be reopened for researchers that are
interested in the Anti-UAV challenge. The benchmark dataset and other information can be found at: https://anti-uav.github.io/.
\end{abstract}


\section{Introduction}

\begin{figure}[h] 
\vspace{-2mm}
\begin{center}
\includegraphics[width=0.85\linewidth]{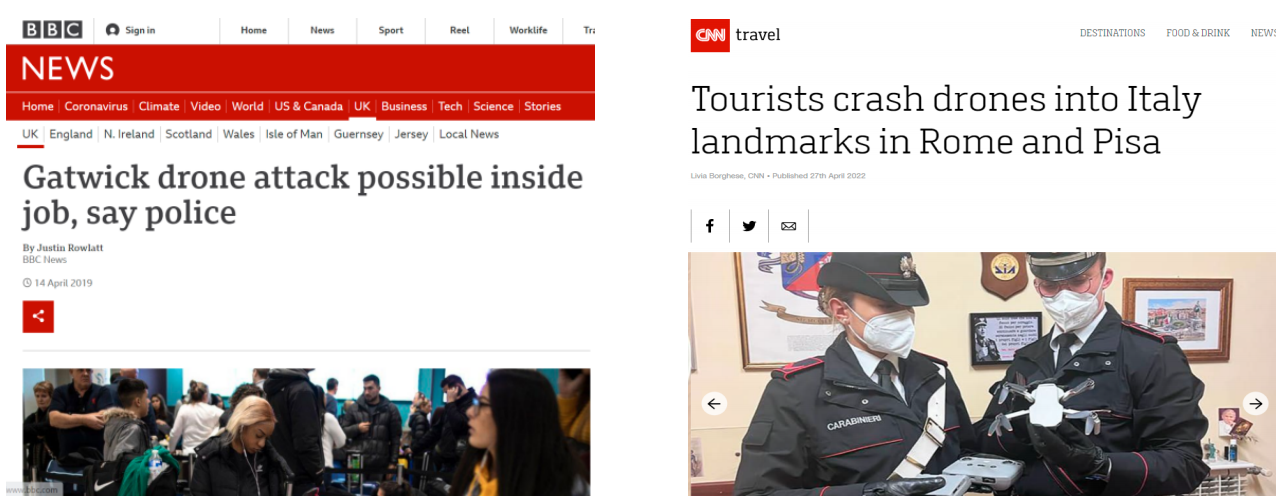}
\end{center}
   \caption{Examples of UAV-related incidents.}
\label{fig:1}
\end{figure}

Civil unmanned aerial vehicles (UAVs), a.k.a. drones, have been widely used in a broad range of civil application domains, including consumer communications, delivery of goods, and remote sensing, owning to their autonomy, flexibility, affordability, and popularity. UAV applications offer possible civil and public domain applications in which single or multiple UAVs may be used. Nevertheless, we should be aware of the potential threat to our lives caused by UAV intrusion since UAVs can also be used to conduct physical attacks ($e.g.$, via explosives) and cyber-attacks ($e.g.$, hacking critical infrastructure). Moreover, unauthorized UAVs sometimes violate aviation safety regulations, thereby bringing hazards to civilian aircraft and passengers and even causing airport disruptions and flight delays. As shown in Fig.~\ref{fig:1}, there have been multiple instances of drone sightings halted air traffic at airports, leading to significant economic losses for airlines. It is highly desired to develop anti-UAV techniques to defend against drone accidents.

Historically, radar is certainly a very powerful technology for detecting traditional incoming airborne threats. However, these comparatively small drones are difficult for radar to accurately detect, because they have very small radar cross-sections and erratic flight paths. Therefore, how to use computer vision algorithms to perceive UAVs is a crucial part of the whole UAV-defense system.

\begin{figure}[h] 
\vspace{-2mm}
\begin{center}
\includegraphics[width=0.85\linewidth]{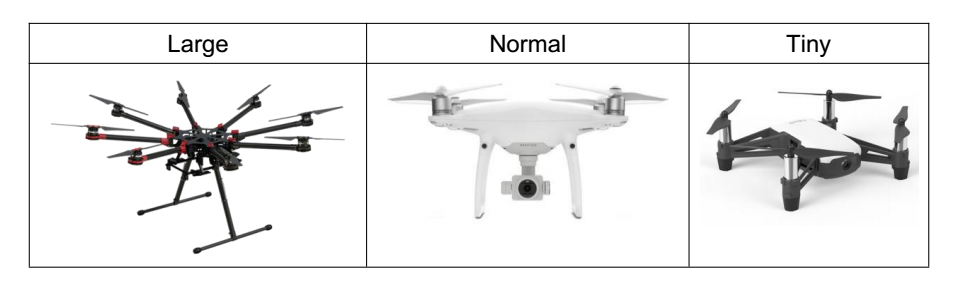}
\end{center}
   \caption{Illustrations of civil UAVs: Large civil UAV; Small civil UAV; Tiny civil UAV.}
\label{fig:2}
\end{figure}

Traditional computer vision research~\cite{Marginalized CNN, Dynamic Conditional Networks,re10,re11,re12} for UAV detection and tracking lacks a high-quality benchmark in dynamic environments. To mitigate this gap, we held the 1st International Workshop on Anti-UAV Challenge~\cite{re13} at CVPR 2020, releasing a dataset consisting of 160 video sequences (both RGB and infrared). The workshop attracted attention from researchers all over the world. Many submitted solutions outperform the baseline method, making great contributions to addressing the anti-UAV problem~\cite{re13,re14,re15}. The 2nd anti-UAV Workshop \& Challenge with ICCV 2021 extends the benchmark dataset to 280 high-quality, full HD thermal infrared video sequences, spanning multiple occurrences of multi-scale ($i.e.$, large, small and tiny, as shown in Fig.~\ref{fig:2}) UAVs. The workshop encourages participants to develop automated methods that can detect and track UAVs in thermal infrared videos with high accuracy. Particularly, algorithms that can detect and track fast-moving drones in complex environments ($e.g.$, occlusion by cloud/buildings/trees, and fake targets like kites, balloons, birds, etc.) are highly expected. The 3rd anti-UAV Workshop \& Challenge for the first time releases the training set and sets two tracks for participants.

This workshop will bring together academic and industrial experts in the field of UAVs to discuss the techniques and applications of tracking UAVs. Participants are invited to submit their original contributions, surveys, and case studies that address the works of UAV’s detection and tracking issues.

This year's challenge has two independent tracks.
\begin{itemize}
    \item \textbf{Track1:} Anti-UAV Tracking Given the bounding box of a drone target in the first frame, this challenge track requires algorithms to track the given target in each video frame by predicting its bounding box. When the target disappears, an invisible mark (no bounding box) needs to be given.
    \item  \textbf{Track2:} Anti-UAV Detection \& Tracking Whether a drone target exits in the first frame is unknown. This challenge track requires algorithms to detect and track the drone target when it appears by predicting its bounding box. When the target does not exist or disappears, an invisible mark (no bounding box) needs to be given.
\end{itemize}

\section{The ANTI-UAV Challenge}
\subsection{Dataset}
The full dataset has been released. There are three subsets in the dataset, $i.e.$, the train subset, the test subset for track 1 and the test subset for track 2. The train subset consists of 200 thermal infrared video sequences and publishes detailed annotation files (whether the target exists, target location information and various challenges). The subset for track 1 also contains 200 video sequences, only providing the position information of target in the first frame; The subset for track 2 contains 200 video sequences. This track does not provide any labeled information. It requires participants to obtain the flag of existence and corresponding target location information of the target through detection and tracking. Above three subsets do not have any overlap between each other. We hope that participants could train a suitable detector or tracker model depending on multiple categories of label information in train subset.

\subsection{Metric}
Anti-UAV is annotated with bounding boxes, attributes and
existing flags. Moreover, an empty bounding box list denotes a ”not exist” flag. Trackers need to obtain the perception of UAV status. In this case, the presence of UAV in the visual range is introduced into the evaluation metric:
\begin{equation}
a c c=\sum_{t=1}^T \frac{I o U_t \times \delta\left(v_t>0\right)+p_t \times\left(1-\delta\left(v_t>0\right)\right)}{T}-0.2 \times\left(\sum_{t=1}^{T^*} \frac{p_t \times \delta\left(v_t>0\right)}{T^*}\right)^{0.3}
\end{equation}
For frame $t$, $IoU_t$ is Intersection over Union (IoU) between the predicted tracking box and its corresponding ground-truth box, $p_t$ is the predicted visibility flag, it equals 1 when the predicted box is empty and 0 otherwise. The $v_t$ is the ground-truth visibility flag of the target, the indicator function $\delta(vt>0)$ equals 1 when $v_t$ > 0 and 0 otherwise. The accuracy is averaged over all frames in a sequence, $T$ indicates total frames and $T*$ denotes the number of frames corresponding to the presence of the target in the ground-truth.

\section{Result and Method}
The 3nd Anti-UAV challenge was held between Feb 6, 2023 and March 13, 2023. The results of the 3nd Anti-UAV challenge are shown in Table 1 and Table 2. Around 20 teams submitted their final results in this challenge. In this section,we will briefly introduce the methodologies of the top 3 submissions of each track.

\subsection{Methods and Experiments in the Track 1}
\subsubsection{Team Colalab}
\textbf{Zongheng Tang, YuluGao, ZiZheng Xun, Fengguang Peng, Yifan Sun, Si Liu, Bo Li.} (Beihang University, Beijing, China \& Baidu Inc.)

The authors adopted the tracking-by-detection strategy to replace the original single tracker. This strategy comprises two main modules, namely, the Strong Detector and the Simple Tracker. They have chosen several types of detectors and optimized each one for the unique characteristics of the UAV object resulting in our Strong Detector. The Simple Tracker uses cascading rules that link the results of the Strong Detector to achieve the final tracking results. However, in infrared images, noise blocks in the background can be similar to the foreground target, resulting in a high probability of detection or tracking failure when relying solely on pure detection methods. To further improve the accuracy of the model, they have utilized temporal information and designed two modules: Video Checker and Motion Model. The Video Checker is a video classifier based on detection results that enlarges the object in the image and crops a local video segment from current and past frames. The segment is then input into the Video Checker for classification, resulting in a new score for the current detection result. The Motion Model is based on background modeling using the frame difference method, which is effective in detecting moving targets with small pixels and can complement detection tasks when dealing with small targets and multiple background clutter similar to foreground targets.

\textbf{Ablation study.} Model Tracking Accuracy.
\begin{enumerate}[-,nosep]
    \item Baseline -> 0.532  
    \item +Threshold -> 0.557 
    \item +Simple Track -> 0.599 
    \item +Motion Model -> 0.607 
    \item +SOT Model -> 0.609
\end{enumerate}

\textbf{Main contribution.} The movement of UAVs can be complex leading to difficulties in real-world scenarios. To address these challenges, they proposes a detection-based method with cascading post-processing modules to solve this task. Their entire process includes generating detection candidate boxes, adjusting candidate box scores through video classification, connecting candidate boxes between different frames through a simple tracker, and determining moving targets in the video through background modeling, followed by single-object tracking as post-processing to adjust the results. This strategy works well for tracking drones of different sizes that are constantly and rapidly moving.

\begin{table}[t]
\centering
\caption{Results of Track 1}
\label{tab1}
\begin{tabular}{
>{\columncolor[HTML]{FFFFFF}}c 
>{\columncolor[HTML]{FFFFFF}}c 
>{\columncolor[HTML]{FFFFFF}}c }
\hline
{\color[HTML]{333333} \textbf{Rank}} & {\color[HTML]{333333} \textbf{User Name}}             & {\color[HTML]{333333} \textbf{Tracking Accuracy}} \\ \hline
{\color[HTML]{333333} 1}             & {\color[HTML]{333333} Colalab(tzhhhh123)}             & {\color[HTML]{333333} 0.700}                     \\
{\color[HTML]{333333} 2}             & {\color[HTML]{333333} USTC-ANTI-UAV(Undefined)}                   & {\color[HTML]{333333} 0.688}                     \\
{\color[HTML]{333333} 3}             & {\color[HTML]{333333} SIA-DT(SIA\_Ryu)}                   & {\color[HTML]{333333} 0.680}                     \\
{\color[HTML]{333333} 4}             & {\color[HTML]{333333} zsl}                 & {\color[HTML]{333333} 0.678}                     \\
{\color[HTML]{333333} 5}             & {\color[HTML]{333333} soro}             & {\color[HTML]{333333}0.677 }                     \\
{\color[HTML]{333333} 6}             & {\color[HTML]{333333} shan666}                       & {\color[HTML]{333333} 0.671}                     \\
{\color[HTML]{333333} 7}             & {\color[HTML]{333333} stephenx24}                    & {\color[HTML]{333333} 0.671}                     \\
{\color[HTML]{333333} 8}             & {\color[HTML]{333333} Silverfall}                & {\color[HTML]{333333} 0.670}                     \\
{\color[HTML]{333333} 9}             & {\color[HTML]{333333} shubo-nlpr}                   & {\color[HTML]{333333}0.667 }                     \\
{\color[HTML]{333333} 10}            & {\color[HTML]{333333} MinkiSong} & {\color[HTML]{333333}0.667 }                     \\ \hline
\end{tabular}
\end{table}

\begin{table}[]
\centering
\caption{Results of Track 2}
\label{tab1}
\begin{tabular}{
>{\columncolor[HTML]{FFFFFF}}c 
>{\columncolor[HTML]{FFFFFF}}c 
>{\columncolor[HTML]{FFFFFF}}c }
\hline
{\color[HTML]{333333} \textbf{Rank}} & {\color[HTML]{333333} \textbf{User Name}}             & {\color[HTML]{333333} \textbf{Tracking Accuracy}} \\ \hline
{\color[HTML]{333333} 1}             & {\color[HTML]{333333} Z-Y}             & {\color[HTML]{333333} 0.611}                     \\
{\color[HTML]{333333} 2}             & {\color[HTML]{333333} FudanDML(ryanhe312)}                   & {\color[HTML]{333333} 0.591}                     \\
{\color[HTML]{333333} 2}             & {\color[HTML]{333333} Colalab(tzhhhh123)}                   & {\color[HTML]{333333}0.570 }                     \\
{\color[HTML]{333333} 4}             & {\color[HTML]{333333} shan666}                 & {\color[HTML]{333333}0.562 }                     \\
{\color[HTML]{333333} 5}             & {\color[HTML]{333333} stephenx24}             & {\color[HTML]{333333} 0.561}                     \\
{\color[HTML]{333333} 6}             & {\color[HTML]{333333}HIT\_HH }                       & {\color[HTML]{333333}0.550 }                     \\
{\color[HTML]{333333} 7}             & {\color[HTML]{333333}shubo-nlpr }                    & {\color[HTML]{333333} 0.540}                     \\
{\color[HTML]{333333} 8}             & {\color[HTML]{333333} KKKKKK}                & {\color[HTML]{333333} 0.538}                     \\
{\color[HTML]{333333} 9}             & {\color[HTML]{333333} QJY0310}                   & {\color[HTML]{333333}0.538 }                     \\
{\color[HTML]{333333} 10}            & {\color[HTML]{333333}Carl\_Huang }            & {\color[HTML]{333333}0.536 }                     \\ \hline
\end{tabular}
\end{table}

\subsubsection{Team USTC-ANTI-UAV}
\textbf{Yinchao Ma, Qianjin Yu, Dawei Yang, Jianfeng He, Yuyang Tang, Tianzhu Zhang} (University of Science and Technology of China)

Authors propose a Unified Transformer-based Tracker,
dubbed UTTracker, which contains four modules, including
background alignment, global detection, multi-region local
tracking and dynamic target detection.

\textbf{Ablation study.} 
\begin{enumerate}[-,nosep]
    \item Base Tracker -> 53.7
    \item Base Tracker + BA -> 57.5
    \item Base Tracker + BA + GD -> 65.9
    \item Base Tracker + BA + GD + MRLT -> 68.3
    \item Base Tracker + BA + GD + MRLT + DTD -> 68.8
\end{enumerate}

\textbf{Main contribution.}
(1) Authors propose a novel Unified Transformer-based Tracker (UTTracker) for robust
UAV tracking, which integrates four modules, including
multi-region local tracking, global detection, background
correction and dynamic small object detection. (2) With
the combine of MRLT, GD and BC modules, our tracker
can achieve robust tracking in challenging scenarios. (3)
To track small target in comlex backgrounds, authors design a improved statistical clustering algorithm to capture
the small UAVs.

\subsubsection{Team SIA-DT}
\textbf{Jeongwon Ryu, Minki Song, Sohee Son.} (SI Analytics)

After visualizing the predictions of the models, we found that FN rarely occurs in drones above the middle, and FP rarely occurs in frames where drones do not exist, according to the coco metric. However, a significant number of FNs occur in drones that are smaller than 10x10 in size, while many FPs and FNs occur due to surrounding objects and camera motion. To address these issues, we employed ensemble and a low score threshold to improve the recall performance for small drones and camouflaged drones caused by surrounding objects and camera motion.

\textbf{Implementation details. }The detector used in this study was configured with the following experimental settings. For the training process, The learning rate was set to 1e-5 with AdamW optimizer and the base anchor scale was changed from 8 to 2. In terms of data augmentation, We followed ms coco augmentation. The backbone architecture used was swin-tiny, which was pre-trained on the ImageNet dataset. Object detector were used: cascading R-CNN, Guided Anchor, Faster R-CNN. In the training process, For YOLOX-X, The learning rate was set to 0.01 / 64 with SGD. Data augmentation added mosaic and mixup in addition to coco ms augmentation. The backbone architecture of YOLOX used modified CSP v5. To make predictions, we set the maximum number of detections to 1 and the score threshold to 0.05, except for YOLOX, which used a score threshold of 0.5 for objectness. Then, we selected the epoch with the best performance on the validation set and used it for ensemble. We used a total of 12 weights for ensemble, including Cascade RCNN trained on the train and validation set at epochs 2, 6, 8, and 12; YOLOX trained on the train set at epoch 10; Faster RCNN trained on the train set at epochs 2 and 3, and Faster RCNN trained on the train and validation set at epochs 2 and 3; and Guided Anchors trained on the train and validation set at epochs 2, 3, and 4.

\textbf{Ablation study.} 
\begin{enumerate}[-,nosep]
    \item YOLOX -> 0.623
    \item Guided anchor -> 0.647
    \item Faster RCNN -> 0.653
    \item Cascade RCNN -> 0.661
    \item Box fusion (YOLOX, Guided anchor, Faster RCNN, Cascade RCNN) -> 0.667
    \item Box fusion (YOLOX 1 model, Guided anchor 3 models, Faster RCNN 4 models, Cascade RCNN 4 models) -> 0.6799
\end{enumerate}

\textbf{Main Contribution. }For the Anti-UAV Challenge, our team aimed to improve UAV tracking by enhancing detector performance. Our approach utilized an ensemble technique to increase detection accuracy in challenging environments. We focused on improving detection accuracy as it is a crucial component of any tracking system. We followed a tracking-by-detection paradigm and believe that our approach can provide valuable insights for future research in this field. Although we did not have enough time to integrate a tracker into our approach, we acknowledge that doing so could enhance tracking performance. This highlights the importance of continued research and development in this area. Our team achieved 3rd place in the 3rd Anti-UAV Challenge Track 1.
\subsection{Methods and Experiments in the Track 2}
\subsubsection{Team Z-Y}
\textbf{Xin Yang, Gang Wang, Weiming Hu, Jin Gao, Shubo Lin, Liang Li, Kai Gao, Yizheng Wang.} (Academy of Military Sciences \& Institute of Automation, Chinese Academy of Sciences)

The authors propose a motion-guided method for small object detection in infrared videos. In video object detection tasks, spatiotemporal motion information plays an essential role in object searching and locating. They introduce the retinal motion extraction algorithm to estimate the motion intensity in consecutive frames. The motion intensity map is used to enhance the possible region features of the moving object, so as to facilitate the video object detection process. Furthermore, in order to alleviate the noise brought by the dynamic background to the motion strength, we introduced spatial attention in the fusion of the motion information and appearance information to specifically enhance the potential moving area. At the same time, a coordinate attention mechanism is added to the end of the backbone network to improve the positioning ability of small objects.

\textbf{Ablation study.} 

\begin{enumerate}[-,nosep]
    \item YOLOv5s(6.0) -> 53.6
    \item YOLOv6m(0.2.0) + motion-guided -> 55.1
    \item YOLOv5s(6.0) + motion-guided -> 61.3
    \item YOLOv5s(6.0) + motion-guided(spatial attention) + Coordinate Attention -> 61.9
    \item YOLOv5l(6.0) + motion-guided(spatial attention) + Coordinate Attention -> 62.3
\end{enumerate}

\subsubsection{Team FudanDML}
\textbf{Ruian He, Shili Zhou, Ri Cheng, Yuqi Sun, Weimin Tan, Bo Yan.} (Fudan University)

The authors propose a novel learning framework for robust UAV detectors called Difference-based Multi-scale Learning (DML). The difference here refers to frame differences, including motion information from previous frames. Multi-scale refers to multiple spatial and temporal scales. First, the method utilizes the frame difference of multiple previous frames, extracting motion information and blocking background noise. Temporal information is essential in small object detection, especially for infrared drones. Because of the UAV's small size and fast motion, the IR background has significant background noise and more occlusions in the complex background. The frame difference method is the classical algorithm to remove background noise and extract motion information. Using multi-frame frame difference as input can improve UAV detection because it can adapt to different motion amplitudes of UAVs. Second, they also fuse the detection results from multiple spatial-temporal scales for inferencing. Exploiting spatial information is also a critical technique for small object detection. The tiny object has features of low recognition, and huge objects are also hard to detect for detectors with fixed anchors and trained on a dataset full of small objects. During training, the model is trained with randomly spatio-temporal augmented input. For inferencing steps, the model predicts from all kinds of augmented input, does Non-maximum Suppression(NMS) for all anchors, and selects the bounding box with the highest score. They implement the method on the popular detector YOLOv5 and significantly improve the performance

\textbf{Ablation study.} 
The evaluation is performed on the validation set. Time Scale 0 means no frame difference is used, and Time Scale 1 means only one frame difference is used. Time Scale 2 and 3 will run the detection for 2 and 3 frame difference inputs for generating final results. Space Scale 1 uses only the original resolution, which is 640 x 512. And Space Scale 2 use [0.5, 1, 2] resolution ratio for inference, and Space Scale 3 use [0.5, 0.75, 1, 1.5, 2] with [0.75, 1.5] flipped left-right. 
\begin{enumerate}[-,nosep]
    \item Time Scale 0: Space Scale 1(Baseline) -> 0.679, Space Scale 3 -> 0.689, Space Scale 5 -> 0.697
    \item Time Scale 1: Space Scale 1 -> 0.769, Space Scale 3 -> 0.780, Space Scale 5 -> 0.785
    \item Time Scale 2: Space Scale 1 -> 0.782, Space Scale 3 -> 0.795, Space Scale 5 -> 0.799
    \item Time Scale 3: Space Scale 1 -> 0.792, Space Scale 3 -> 0.799, Space Scale 5 -> 0.803

\end{enumerate}

\subsubsection{Team Colalab}
\textbf{Zongheng Tang, YuluGao, ZiZheng Xun, Fengguang Peng, Yifan Sun, Si Liu, Bo Li.} (Beihang University, Beijing, China \& Baidu Inc.)

The authors of this work replaced the original single tracker with a tracking-by-detection strategy, which consists of two main modules: the Strong Detector and the Simple Tracker. They selected several types of detectors and optimized each one for the unique characteristics of the UAV object, resulting in a Strong Detector that is effective at detecting the objects of interest. The Simple Tracker uses cascading rules to link the results of the Strong Detector to achieve the final tracking results. However, in infrared images, noise blocks in the background can be similar to the foreground target, making it difficult to rely solely on pure detection methods. To further improve the accuracy of the model, the authors utilized temporal information and designed two modules: the Video Checker and the Motion Model. The Video Checker is a video classifier based on detection results that enlarges the object in the image and crops a local video segment from current and past frames. The segment is then input into the Video Checker for classification, resulting in a new score for the current detection result. The Motion Model is based on background modeling using the frame difference method, which is effective in detecting moving targets with small pixels and can complement detection tasks when dealing with small targets and multiple background clutter similar to foreground targets. These modules help to overcome the limitations of pure detection methods and improve the accuracy of the tracking system.

\textbf{Ablation study.} Model Tracking Accuracy 

\begin{enumerate}[-,nosep]
    \item Baseline -> 0.532
    \item +Simple Track -> 0.599 
    \item +Motion Model -> 0.607 
\end{enumerate}

\textbf{Main Contribution. }The movement patterns of UAVs can be complex, making it challenging to track them in real-world scenarios. To address these challenges, the authors proposed a detection-based method that incorporates cascading post-processing modules to solve this task. Their approach involves generating detection candidate boxes, adjusting the candidate box scores through video classification, connecting candidate boxes between different frames using a simple tracker, and determining moving targets in the video through background modeling. As a post-processing step, single-object tracking is used to adjust the results.

This strategy has been found to work well for tracking drones of different sizes that are constantly and rapidly moving. By incorporating multiple modules, the proposed method is able to effectively deal with the challenges posed by complex UAV movement patterns, and produce accurate tracking results.

\section{Conclusions}
Object detection and tracking in the wild scenarios are fundamental yet challenging problems in computer vision. We held the 3rd Anti-UAV Challenge to encourage researchers from the fields of object detection, visual tracking and other disciplines to present their progress, communication and novel ideas that will potentially shape the future of the UAV detection area. Approximately 76 teams around the globe participated in this competition, in which top-3 leading teams in each track, together with their methods, are briefly introduced in this paper. Our workshop takes a different perspective, making UAVs as tracking targets, and provides a large-scale dataset to promote deep network learning for UAVs. In addition, the proposed workshop also aims at tiny object detection and tracking in the wild which is more challenging, more practical, and more useful for real applications. Thus, our workshop will bridge the needs of industry and research in academia, and may accelerate the process of these computer vision technologies being used in real applications.

\section{Acknowledgement}
This work is partially supported by National Natural Science Foundation of China (62006244), Young Elite Scientist Sponsorship Program of China Association for Science and Technology (YESS20200140), and Young Elite Scientist Sponsorship Program of Beijing Association for Science and Technology (BYESS2021178).

\bibliographystyle{unsrt}  


\begin{thebibliography}{100}

\bibitem{SiamR-CNN}
Paul Voigtlaender, Jonathon Luiten, Philip H.S. Torr, Bastian Leibe.
\newblock Siam R-CNN: Visual Tracking by Re-Detection.
\newblock In {\em Proceedings of the IEEE/CVF Conference on Computer Vision and Pattern Recognition (CVPR)}, pages 6578-6588. 2020.

\bibitem{ECO}
Martin Danelljan, Goutam Bhat, Fahad Shahbaz Khan, Michael Felsberg.
\newblock ECO: Efficient Convolution Operators for Tracking.
\newblock In {\em Proceedings of the IEEE/CVF Conference on Computer Vision and Pattern Recognition (CVPR)}, pages 6638-6646. 2017.


\bibitem{bhat2019learning}
Fredrik K. Gustafsson, Martin Danelljan, Radu Timofte, Thomas B. Schön.
\newblock How to Train Your Energy-Based Model for Regression.
\newblock {\em arXiv preprint arXiv:2005.01698}, 2020.

\bibitem{LTMU}
Kenan Dai, Yunhua Zhang, Dong Wang, Jianhua Li, Huchuan Lu, Xiaoyun Yang.
\newblock High-Performance Long-Term Tracking With Meta-Updater.
\newblock In {\em Proceedings of the IEEE/CVF Conference on Computer Vision and Pattern Recognition (CVPR)}, pages 6298-6307. 2020.

\bibitem{AlphaRef}
Bin Yan, Dong Wang, Huchuan Lu, Xiaoyun Yang.
\newblock Alpha-Refine: Boosting Tracking Performance by Precise Bounding Box Estimation.
\newblock {\em arXiv preprint arXiv:2007.02024}, 2020.

\bibitem{li2019siamrpn++}
Bo Li, Wei Wu, Qiang Wang, Fangyi Zhang, Junliang Xing, Junjie Yan.
\newblock SiamRPN++: Evolution of Siamese Visual Tracking With Very Deep Networks.
\newblock In {\em Proceedings of the IEEE/CVF Conference on Computer Vision and Pattern Recognition (CVPR)}, pages 4282-4291. 2019.

\bibitem{chen2021transformer}
Xin Chen, Bin Yan, Jiawen Zhu, Dong Wang, Xiaoyun Yang, Huchuan Lu.
\newblock Transformer Tracking.
\newblock In {\em Proceedings of the IEEE/CVF Conference on Computer Vision and Pattern Recognition (CVPR)}, pages 8126-8135. 2021.



\bibitem{Marginalized CNN}
Jian ZHAO, Jianshu Li, Fang Zhao, Xuecheng Nie, Yunpeng Chen, Shuicheng Yan and Jiashi Feng.
\newblock Marginalized CNN: Learning Deep Invariant Representations.
\newblock In {\em Proceedings of the British Machine Vision Conference (BMVC)}, pages 127.1-127.12. 2017.

\bibitem{Dynamic Conditional Networks}
Fang Zhao, Jian Zhao, Shuicheng Yan, Jiashi Feng.
\newblock Dynamic Conditional Networks for Few-Shot Learning.
\newblock In {\em Proceedings of the European Conference on Computer Vision (ECCV)}, pages 19-35. 2018.

\bibitem{re10}
Li Zhou, Jian Zhao, Jianshu Li, Li Yuan, Jiashi Feng.
\newblock Object Relation Detection Based on One-shot Learning.
\newblock In {\em Proceedings of the IEEE/CVF Conference on Computer Vision and Pattern Recognition (CVPR)}. 2018.

\bibitem{re11}
Jianshu Li, Pan Zhou, Yunpeng Chen, Jian Zhao, Sujoy Roy, Yan Shuicheng, Jiashi Feng and Terence Sim.
\newblock Task Relation Networks.
\newblock In {\em 2019 IEEE Winter Conference on Applications of Computer Vision (WACV)}, pages 932-940. 2019.

\bibitem{re12}
Yuqi Gong, Xuehui Yu, Yao Ding, Xiaoke Peng, Jian Zhao, Zhenjun Han.
\newblock Effective Fusion Factor in FPN for Tiny Object Detection.
\newblock In {\em Proceedings of the IEEE/CVF Conference on Computer Vision and Pattern Recognition (CVPR)}. 2020.

\bibitem{re13}
Nan Jiang, Kuiran Wang, Xiaoke Peng, Xuehui Yu, Qiang Wang, Junliang Xing, Guorong Li, Jian Zhao, Guodong Guo and Zhenjun Han.
\newblock Anti-UAV: A Large Multi-Modal Benchmark for UAV Tracking.
\newblock In {\em Proceedings of the IEEE/CVF Conference on Computer Vision and Pattern Recognition (CVPR)}. 2021.

\bibitem{re14}
Liqian Liang, Congyan Lang, Zun Li, Jian Zhao, Tao Wang, and Songhe Feng.
\newblock Seeing Crucial Parts: Vehicle Model Verification via A Discriminative Representation Model.
\newblock In {\em Journal of the ACM (JACM)}, pages 22. 2018.

\bibitem{re15}
Zun Li, Congyan Lang, Liqian Liang, Jian Zhao, Songhe Feng, Qibin Hou and Jiashi Feng.
\newblock Dense Attentive Feature Enhancement for Salient Object Detection.
\newblock In {\em IEEE Transactions on Circuits and Systems for Video Technology}. 2021.




\end{thebibliography}

\vspace*{3\baselineskip}

\begin{figwindownonum}[0,l,{\mbox{\includegraphics[width=3cm]{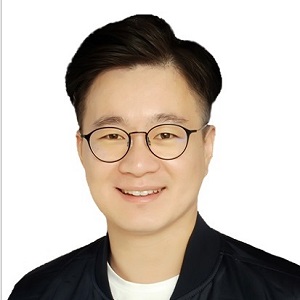}}},{}]
\noindent
\textbf{Jian Zhao} is with Institute of North Electronic Equipment, Beijing, China; Intelligent Game and Decision Laboratory, Beijing, China; Department of Mathematics and Theories, Peng Cheng Laboratory, Shenzhen, China. He received his Ph.D. degree from National University of Singapore (NUS) in 2019 under the supervision of Assist. Prof. Jiashi Feng and Assoc. Prof. Shuicheng Yan. He is the SAC of VALSE, the committee member of CSIG-BVD, and the member of the board of directors of BSIG. He has received the "2020-2022 Youth Talent Promotion Project" from China Association for Science and Technology, and the "2021-2023 Beijing Youth Talent Promotion Project" from Beijing Association for Science and Technology. He has published over 40 cutting-edge papers on human-centric image understanding. He has won the Lee Hwee Kuan Award (Gold Award) on PREMIA 2019 and the "Best Student Paper Award" on ACM MM 2018 as the first author. He has received the nomination for the USERN Prize 2021, according to publications as first author in top rank (Q1) journals of the field of Artificial Intelligence, Pattern Recognition, Machine Learning, Computer Vision and Multimedia Analytics, in the recent two years. He has won the top-3 awards several times on world-wide competitions on face recognition, human parsing and pose estimation as the first author. His main research interests include deep learning, pattern recognition, computer vision and multimedia. He and his collaborators has also successfully organized the CVPR 2020 Anti-UAV Workshop and Challenge, the ECCV 2020 RLQ-TOD Workshop and Challenge, and the CVPR 2018 L.I.P Workshop and MHP Challenge.
\end{figwindownonum}

\begin{figwindownonum}[0,l,{\mbox{\includegraphics[width=3cm]{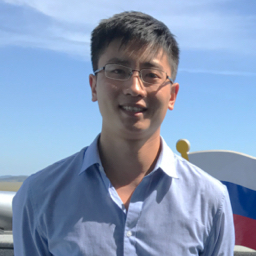}}},{}]
\noindent
\textbf{Jianan Li} is currently an Assistant Professor at School of Optoelectronics, Beijing Institute of Technology, Beijing, China, where he received his B.S. and Ph.D. degree in 2013 and 2019, respectively. From July 2019 to July 2020, he worked as a research fellow at National University of Singapore, where he also worked as a joint training Ph.D. student from July 2015 to July 2017. From October 2017 to April 2018, he worked as an intern at Adobe Research. His research interests mainly include computer vision and real-time image/video processing.
\end{figwindownonum}
\vspace{0.5 cm}

\begin{figwindownonum}[0,l,{\mbox{\includegraphics[width=3cm]{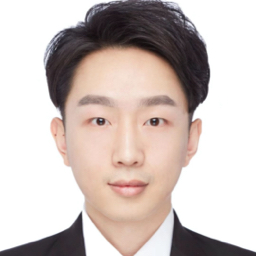}}},{}]
\noindent
\textbf{Lei Jin} is currently a Postdoc with the Beijing University of Posts and Telecommunications (BUPT), Beijing, China. He graduated from the same university with a
Ph.D. degree in 2020. He received the bachelor degree in the BUPT in 2015. His research
interests include network security, traffic security analysis, machine learning and pattern recognition, with a focus on 6Dof poses estimation and Human pose estimation.
\end{figwindownonum}
\vspace{3 cm}

\begin{figwindownonum}[0,l,{\mbox{\includegraphics[width=3cm]{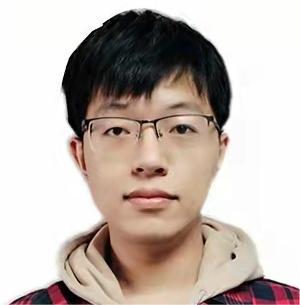}}},{}]
\noindent
\textbf{Jiaming Chu} is currently pursuing a doctoral degree in Electronic Science and Technology at Beijing University of Posts and Telecommunications. His research interests include deep learning and computer vision.
\end{figwindownonum}
\vspace{2 cm}

\begin{figwindownonum}[0,l,{\mbox{\includegraphics[width=3cm]{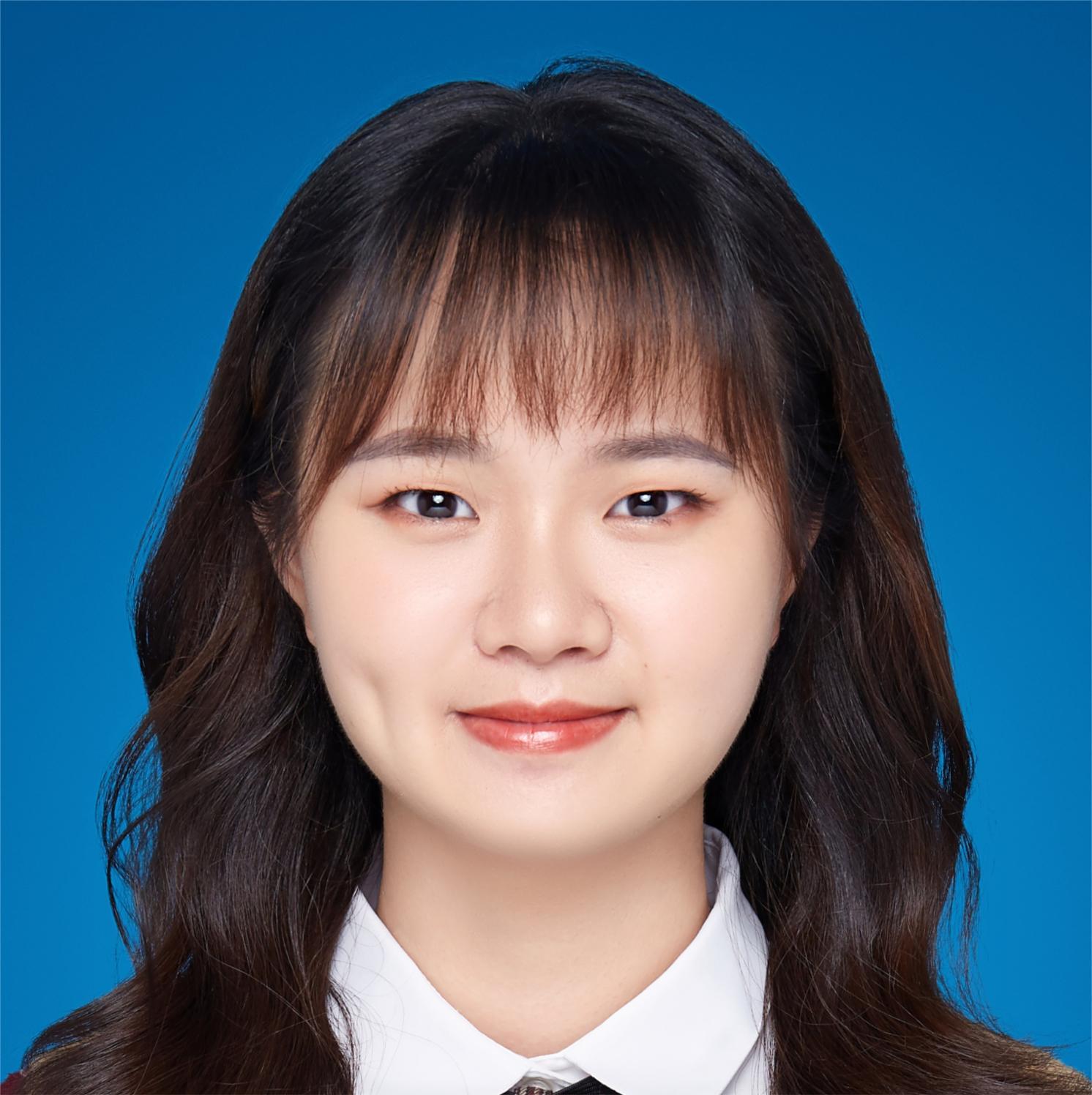}}},{}]
\noindent
\textbf{Jun Wang} is a postgraduate student at National Defense Innovation Institute.
Her main research interests include computer vision, object tracking.
\end{figwindownonum}
\vspace{2.2 cm}

\begin{figwindownonum}[0,l,{\mbox{\includegraphics[width=3cm]{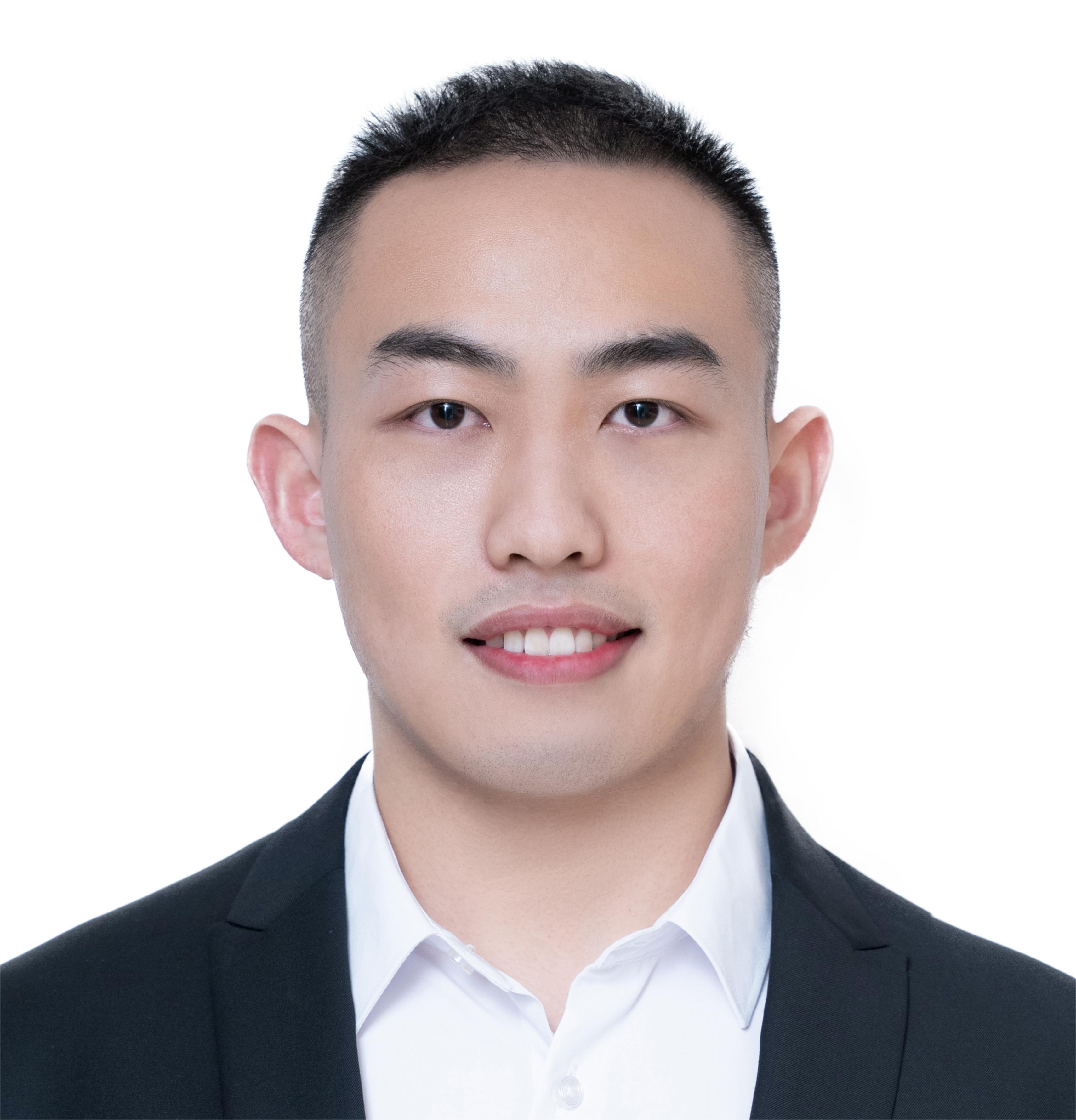}}},{}]
\noindent
\textbf{Jianqiang Xia} is currently pursuing a master's degree at the National Innovation Institute of Defense Technology, Beijing, China. His research interests mainly focus on object detection/tracking. He won the first prize in the 4th "Huawei Cup" China Graduate Artificial Intelligence Innovation Competition in 2022.
\end{figwindownonum}
\vspace{1.7 cm}

\begin{figwindownonum}[0,l,{\mbox{\includegraphics[width=3cm]{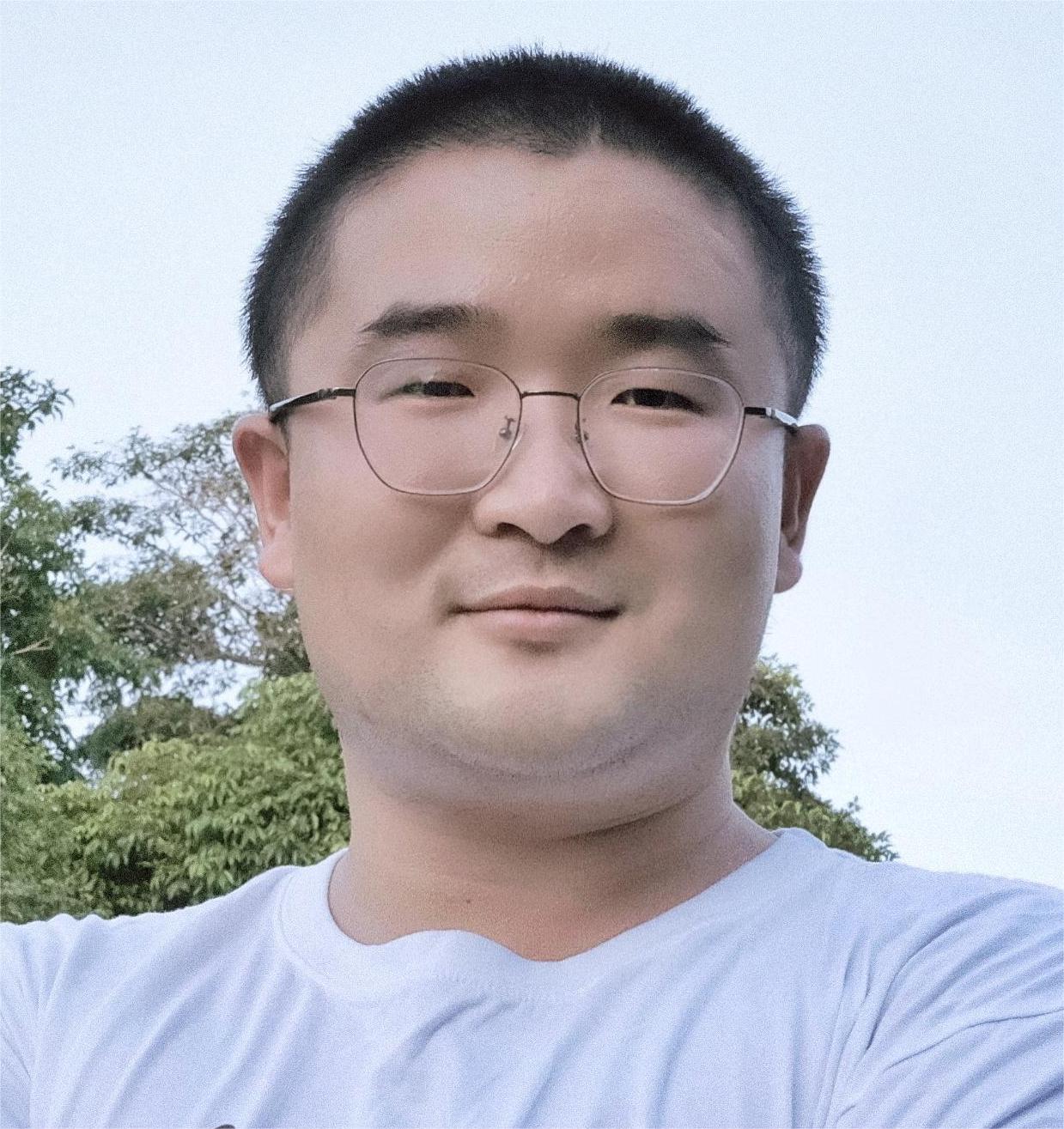}}},{}]
\noindent
\textbf{Kai Wang} is currently a Ph.D. student at National University of Singapore, supervised 
by Prof. Yang You. Before that, he received the master degree from University of Chinese 
Academy of Sciences in 2020. During 2020-2021, he was research interns at Huawei and 
Alibaba Damo Academy. His research areas include Resource Efficient AI, Affective 
Computing and Noise Robust Learning. 
\end{figwindownonum}
\vspace{1.5 cm}

\begin{figwindownonum}[0,l,{\mbox{\includegraphics[width=3cm]{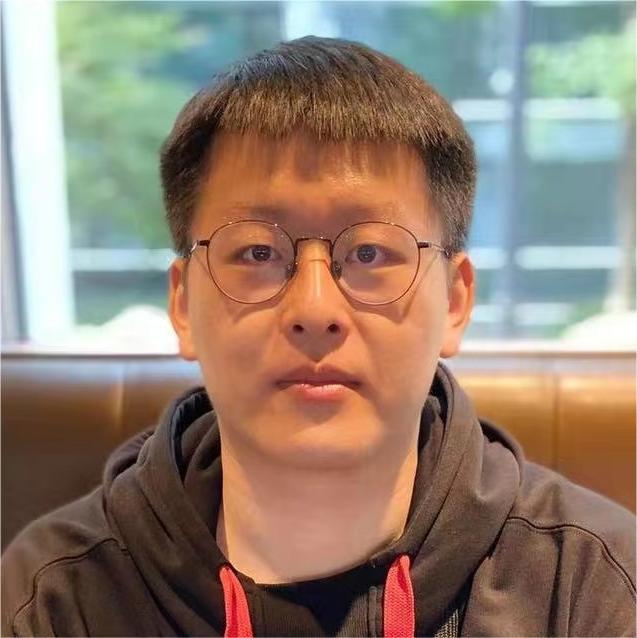}}},{}]
\noindent
\textbf{Yang Liu} is a senior researcher in the DAMO Machine Intelligence Laboratory at Institute of Alibaba.
He received his M.Sc. degrees  in automation from North China Electric Power University, China in 2020. 
He is a reviewer in prestigious computer vision conferences including CVPR, ICCV, ICML, ICLR and NeurIPS.
His main research interests include face detection/recognition, zero-shot learning, unsupervised domain adaptation.
\end{figwindownonum}
\vspace{1.5 cm}

\begin{figwindownonum}[0,l,{\mbox{\includegraphics[width=3cm]{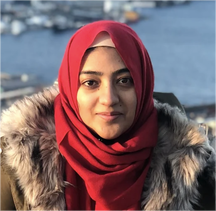}}},{}]
\noindent
\textbf{Sadaf Gulshad} is currently a Postdoc in the Multix Lab, University of Amsterdam, the 
Netherlands. Previously, she did her PhD in ML and CV from Bosch Delta Lab, University 
of Amsterdam. Before starting her PhD she completed her Masters in Electrical Engineering 
from KAIST, South Korea. Her current research interests lie in understanding and explaining 
the decisions of neural networks for video action classification. Previously she worked on 
robustifying neural networks against adversarial and natural perturbations.
\end{figwindownonum}
\vspace{0.8 cm}

\begin{figwindownonum}[0,l,{\mbox{\includegraphics[width=3cm]{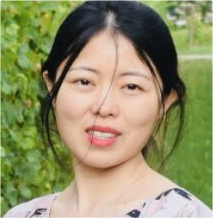}}},{}]
\noindent
\textbf{Jiaojiao Zhao} obtained her PhD degree in the VIS Lab, University of Amsterdam, the 
Netherlands. Before her PhD, she worked as a research assistant in the Artificial Intelligence 
Lab, University of East Anglia, UK. And from 
 2015 to 2016, she worked in Panasonic R\&D 
 Center Singapore and visited the Learning and Vision Group at National University of 
Singapore. Her current research interests are video analysis including human action detection, 
tracking and recognition, and machine learning.
\end{figwindownonum}
\vspace{1 cm}

\begin{figwindownonum}[0,l,{\mbox{\includegraphics[width=3cm]{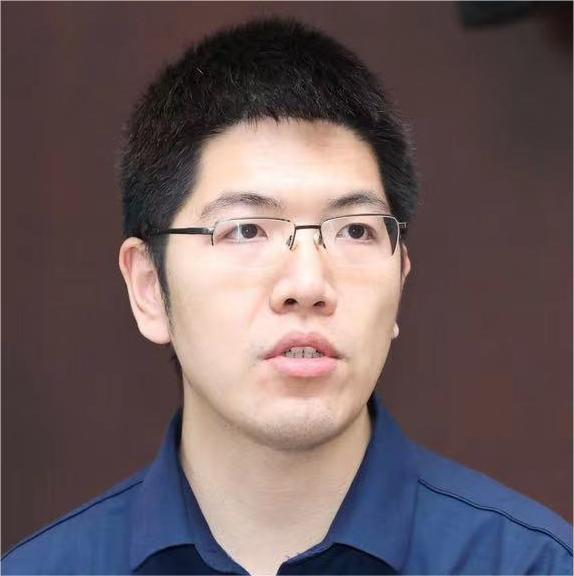}}},{}]
\noindent
\textbf{Tianyang Xu} is currently an Associate Professor at the School of Artificial Intelligence 
and Computer Science, Jiangnan University, Wuxi, China. He received the B.Sc. degree in 
electronic science and engineering from Nanjing University, Nanjing, China, in 2011. He 
received the PhD degree at the School of Artificial Intelligence and Computer Science, 
Jiangnan University, Wuxi, China, in 2019. He was a Research Fellow in CVSSP, University 
of Surrey during 2019 to 2021. His research interests include visual tracking and deep 
learning. He has published several scientific papers, including IJCV, ICCV, TIP, TIFS, TKDE, 
etc. He achieved top 1 performance in academic competitions, including the VOT2018 public 
dataset (ECCV18), VOT2020 RGBT challenge (ECCV20), Anti-UAV challenge (CVPR20), 
Multi-Modal Video Reasoning and Analysing Competition (ICCV21).
\end{figwindownonum}
\vspace{0.2 cm}

\begin{figwindownonum}[0,l,{\mbox{\includegraphics[width=3cm]{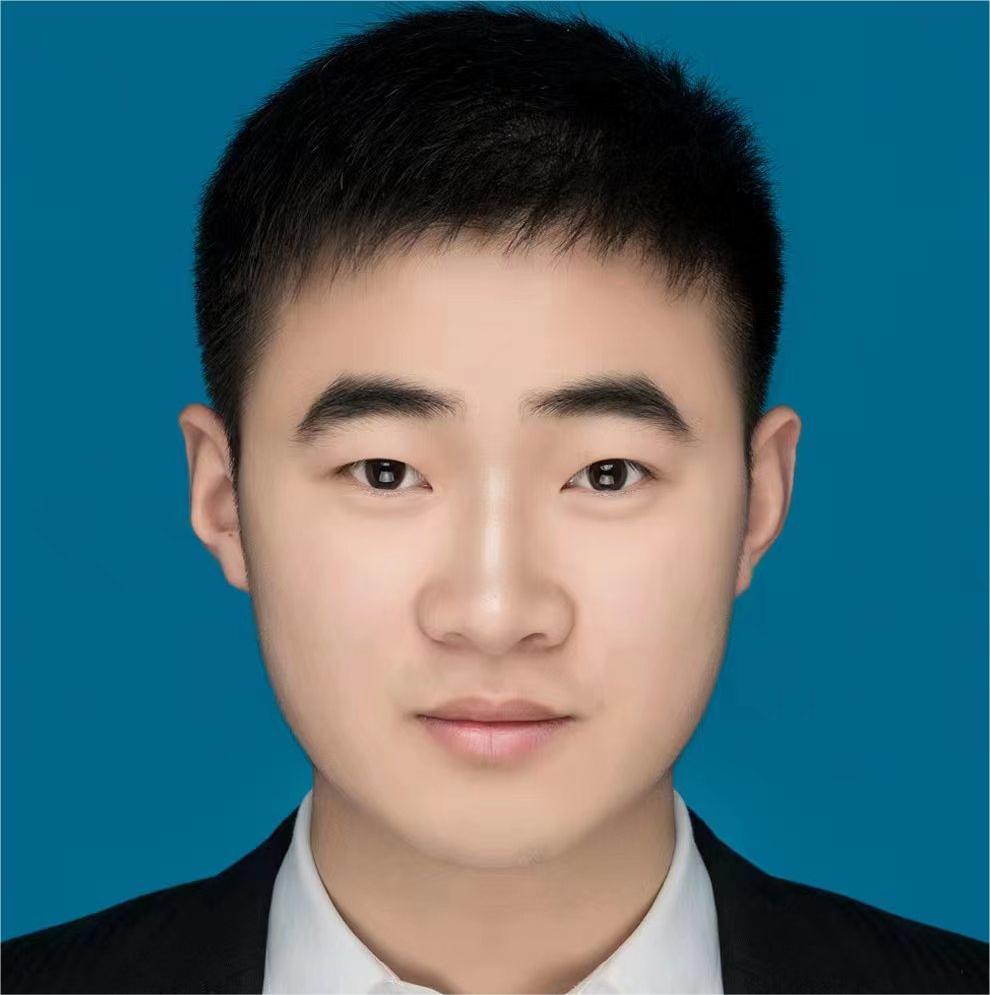}}},{}]
\noindent
\textbf{Xue-Feng Zhu} received the B.Eng. degree in Internet of Things Engineering from Chongqing University of Posts and Telecommunications, Chongqing, China, in 2017. Currently, he is a PhD student at the School of Artificial Intelligence and Computer Science, Jiangnan University, Wuxi, China. He is also a visiting PhD student with the Centre for Vision, Speech and Signal Processing (CVSSP), University of Surrey, Guildford, U.K. His research interests include visual object tracking and multi-modal learning. He has published several scientific papers, including 2023 AAAI Conference, IEEE Transactions on Multimedia, IEEE Transactions on Circuits and Systems for Video Technology, Pattern Recognition etc. He achieved the 1st place award in the VOT2020 RGBT challenge (ECCV2020) and the 3rd place award in the 2nd Anti-UAV challenge (CVPR2021).
\end{figwindownonum}
\vspace{0.2 cm}

\begin{figwindownonum}[0,l,{\mbox{\includegraphics[width=3cm]{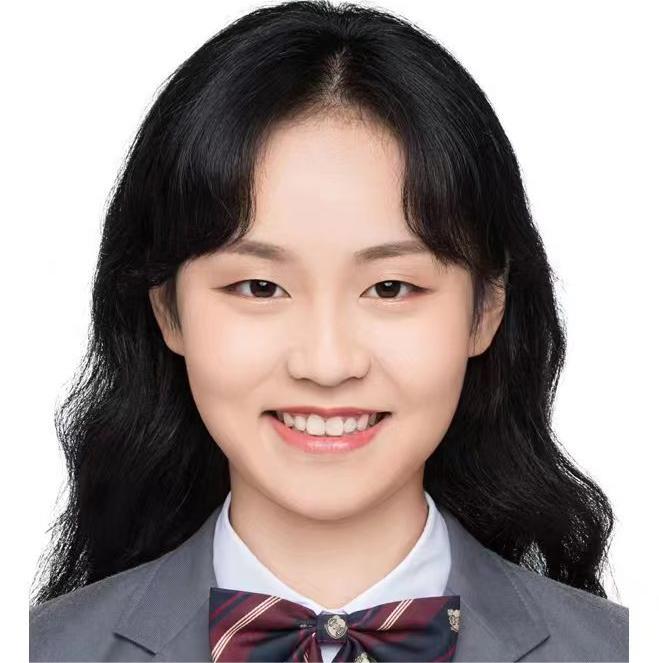}}},{}]
\noindent
\textbf{Shihan Liu} is a postgraduate student in Jiangnan University, She received her bachelor's degree in Internet of Things Engineering from Jiangnan University in 2022. She  has great interest in computer vision especially object tracking and detection.
\end{figwindownonum}
\vspace{2.2 cm}

\begin{figwindownonum}[0,l,{\mbox{\includegraphics[width=3cm]{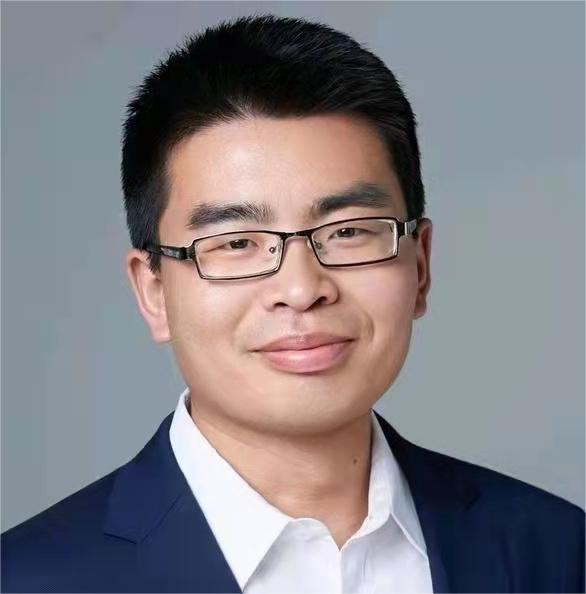}}},{}]
\noindent
\textbf{Zheng Zhu} is currently the Research Director and Scientist at PhiGent Robotics. During 
2019-2021, he was a post-doc fellow at Tsinghua University, worked with Prof. Jiwen Lu. 
He received Ph.D. degree from Institute of Automation, Chinese Academy of Sciences in 
2019. During 2016-2019, he was research interns at SenseTime, Horizon Robotics, and 
DeepGlint. He served reviewers in various journals and conference including IEEE-TPAMI, 
IEEE-TMM, IEEE-TCSVT, CVPR, ICCV, ECCV, ICLR. He has co-authored 40+ journal 
and conference papers mainly on computer vision and robotics problems, such as face 
recognition, visual tracking, human pose estimation, and servo control. He has more than 
3,500 Google Scholar citations to his work, including SiamRPN (1,000+ citations) and 
DaSiamRPN (600+ citations). His work DaSiamRPN is included in famous OpenCV Library. 
He organized the Masked Face Recognition Challenge \& Workshop (MFR) in ICCV 2021. 
He ranked the 1st on NIST-FRVT Masked Face Recognition, won the COCO Keypoint 
Detection Challenge in ECCV 2020 and Visual Object Tracking (VOT) Real-Time Challenge 
in ECCV 2018.
\end{figwindownonum}

\begin{figwindownonum}[0,l,{\mbox{\includegraphics[width=3cm]{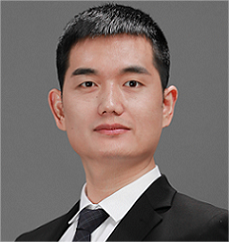}}},{}]
\noindent
\textbf{Guibo Zhu} is currently an Associate Professor with the National Laboratory of Pattern 
Recognition (NLPR), Institute of Automation, Chinese Academy of Sciences (CASIA), 
Beijing, China. He received the B.E. degree from Wuhan University, Wuhan, China, in 2009, 
the Ph.D. degree from the University of Chinese Academy of Sciences, Beijing, China, in 
2016. He has published over 10 papers as the first author in international conferences like 
AAAI, IJCAI, BMVC, and journals like T-IP, T-NNLS, T-Cyber, CVIU. He has won two 
champions of domestic AI technical competitions in single visual tracking and video 
understanding as the first author. His research interests include computer vision, pattern 
recognition, and machine learning, especially paying attention on single object tracking, 
object detection, video understanding and meta-learning.
\end{figwindownonum}
\vspace{0.1 cm}

\begin{figwindownonum}[0,l,{\mbox{\includegraphics[width=3cm]{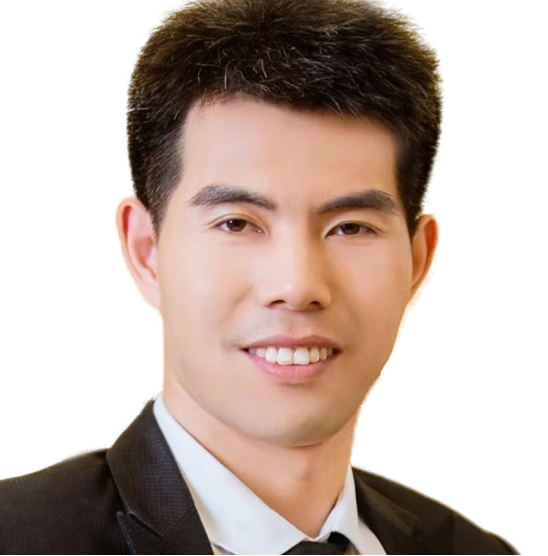}}},{}]
\noindent
\textbf{Zechao Li} is currently a Professor at the Nanjing University of Science and Technology. He received the Bachelor’s degree from University of Science and Technology of China (USTC) in 2008 and the Doctor degree from National Laboratory of Pattern Recognition (NLPR), Institute of Automation, Chinese Academy of Sciences in 2013. He has authored over 90 papers in top-tier journals and conferences. His research interests include multimedia analysis, object detection, semantic segmentation, few-shot learning, etc. He was a recipient of the Best Paper Award in ACM Multimedia in Asia 2020, and the Best Student Paper Award in ICIMCS 2018. He has served as an Associate Editor for IEEE TNNLS, Information Sciences, JCST, etc.
\end{figwindownonum}
\vspace{0.3 cm}

\begin{figwindownonum}[0,l,{\mbox{\includegraphics[width=3cm]{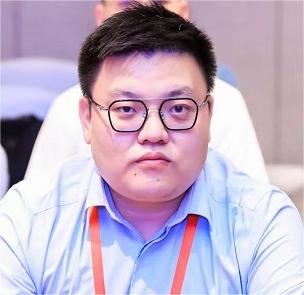}}},{}]
\noindent
\textbf{Zheng Wang} is a Professor with Wuhan University, Wuhan, China. He was a Project 
Researcher and JSPS Fellowship Researcher with the National Institute of Informatics, Japan 
from 2017 to 2020, and a Project Assistant Professor with the Research Institute for an 
Inclusive Society through Engineering (RIISE), the University of Tokyo, in 2021. He had coorganized the ACM MM 2020 Tutorial on “Effective and Efficient: Toward Open-world 
Instance Re-identification”, CVPR 2020 Tutorial on “Image Retrieval in the Wild”, and ACM 
MM 2022 Tutorial on “Multimedia Content Understanding in Harsh Environments”. His 
research interests focus on multimedia content analysis.
\end{figwindownonum}
\vspace{0.2 cm}

\begin{figwindownonum}[0,l,{\mbox{\includegraphics[width=3cm]{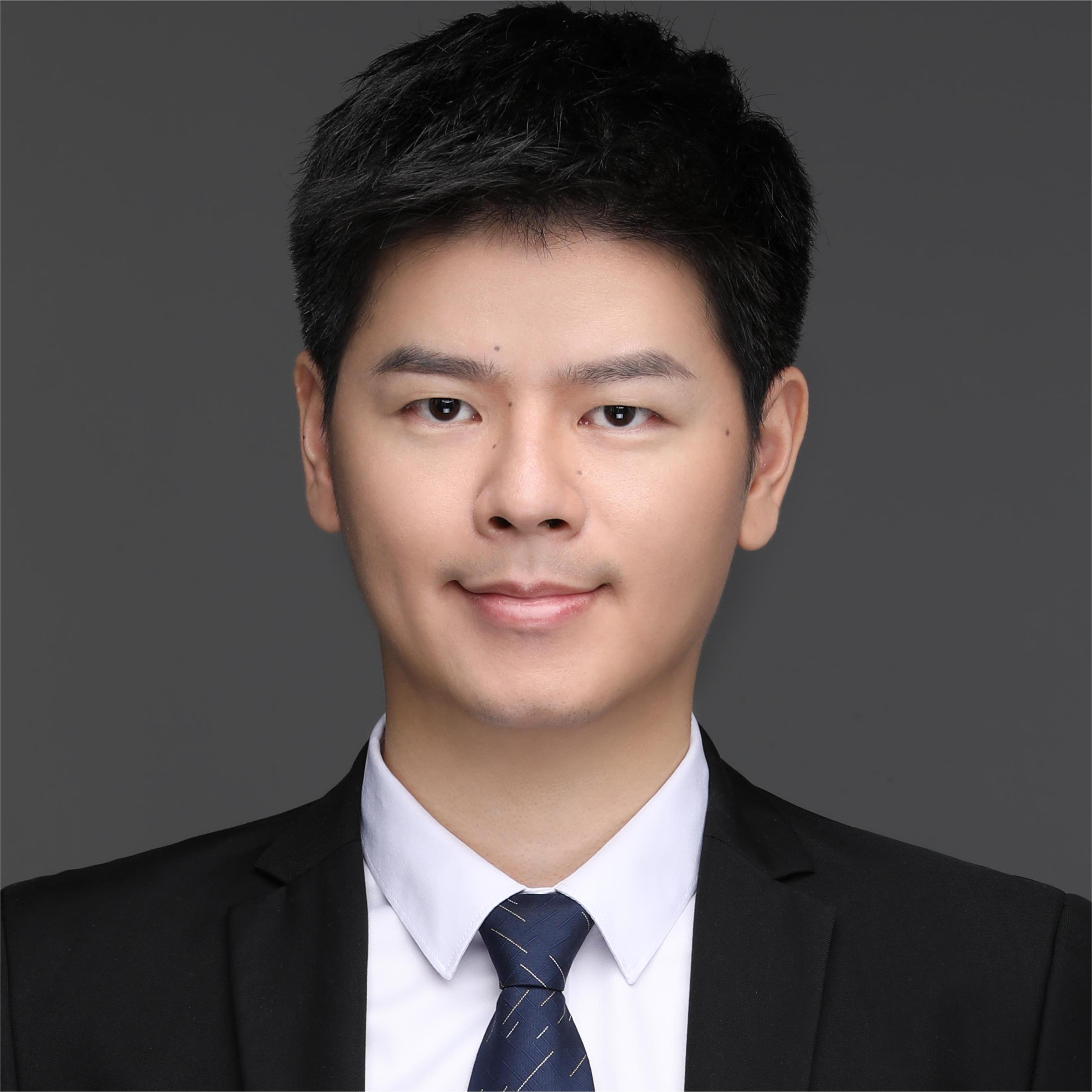}}},{}]
\noindent
\textbf{Baigui Sun} received the M.Eng. degree from Zhejiang University, Hangzhou, China, in 2014. He joined Alibaba Group, Hangzhou, China, as a computer vision algorithm researcher, in 2014. His research interests include face detection/recognition, deep metric learning, domain adaption, zero/few-shot learning, semi-supervised learning and computer vision applications. He has published over 10 articles in international journals and conferences.
\end{figwindownonum}
\vspace{1.5 cm}

\begin{figwindownonum}[0,l,{\mbox{\includegraphics[width=3cm]{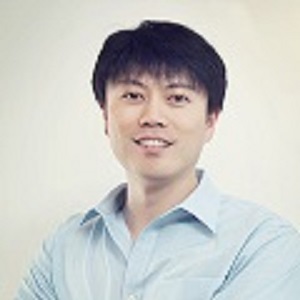}}},{}]
\noindent
\textbf{Yandong Guo} is currently the OPPO Chief Scientist of Intelligent Perception and 
Adjunct Professor of Beijing Univ   ersity of Posts and Telecommunications. He gained his 
PhD from School of Electronic and Computer Engineering, Purdue University (West 
Lafayette) under the supervision of Prof. Jan P. Allebach and Charles A. Bouman. Dr. Guo 
mainly focuses on computer vision and artificial intelligence research and applies the 
research in industry applications. His papers have been widely accepted in CVPR, ECCV, 
ICML and other internationally recognized academic conferences and journals, and have 
been cited thousands of times by peers, and have also empowered many core products of 
companies including GE, HP, Microsoft, Xpeng Motors and OPPO. He has also served as a 
program committee member or reviewer for multiple conferences and journals and has 
organized ICCV and CVPR Workshops as the chair member.
\end{figwindownonum}

\begin{figwindownonum}[0,l,{\mbox{\includegraphics[width=3cm]{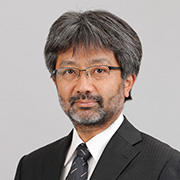}}},{}]
\noindent
\textbf{Shin’ichi Satoh} is a Professor at National Institute of Informatics (NII) and University 
of Tokyo. Satoh is currently the Director of Research Center for Medical Big Data, NII. He 
is a Fellow of IEICE and ITE and has served as Editorial Board/Associate Editor for lots of 
top Journals, such as IJCV, TMM, TCSVT. He has published 4 graduate level textbooks and 
over 250 scientific papers in well recognized journals including TIP, TMM, TCSVT, TCYB, 
and top international conference including ACM MM, CVPR, ICCV, ECCV, AAAI, IJCAI. 
Satoh’s research interests include multimedia information analysis and knowledge discovery, 
aiming to create an intelligent computer system to see and understand the visual worl.
\end{figwindownonum}
\vspace{0.2 cm}

\begin{figwindownonum}[0,l,{\mbox{\includegraphics[width=3cm]{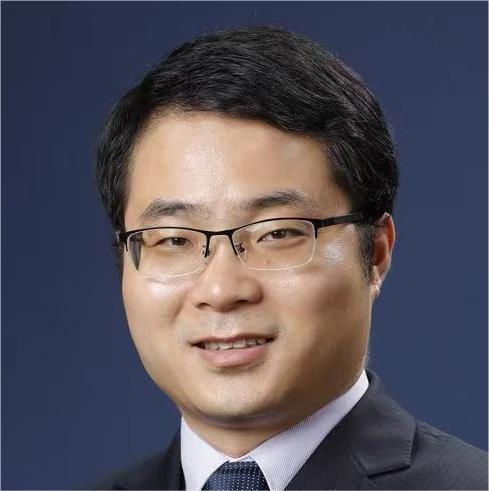}}},{}]
\noindent
\textbf{Junliang Xing} is a Professor with Tsinghua University and the recipient of the National 
Science Fund for Excellent Young Scholar. He obtained his dual bachelor's degrees in 
Computer Science and Mathematics at Xi'an Jiaotong University in 2007 and his doctorate 
in Computer Science in 2012. Then he worked in the National Laboratory of Pattern 
Recognition, Institute of Automation, Chinese Academy of Sciences as an assistant 
researcher, associated researcher, and researcher in 2012, 2015, and 2018, respectively. His 
research interests are human-computer interactive learning, computer gaming, and computer 
vision. He has published more than 100 peer-reviewed papers in international conferences 
and journals and got more than 13,000 citations from Google Scholar. He has also published 
two books and translated three classical textbooks in artificial intelligence. He has been 
granted the "New People in Academia", "Google Fellowship", and three times 
"Best/Distinguished Paper" awards in top international and domestic conferences, with a 
dozen times of prizes in AI technique competitions. His research techniques have got several 
applications in practical systems.
\end{figwindownonum}
\vspace{0.2 cm}

\begin{figwindownonum}[0,l,{\mbox{\includegraphics[width=3cm]{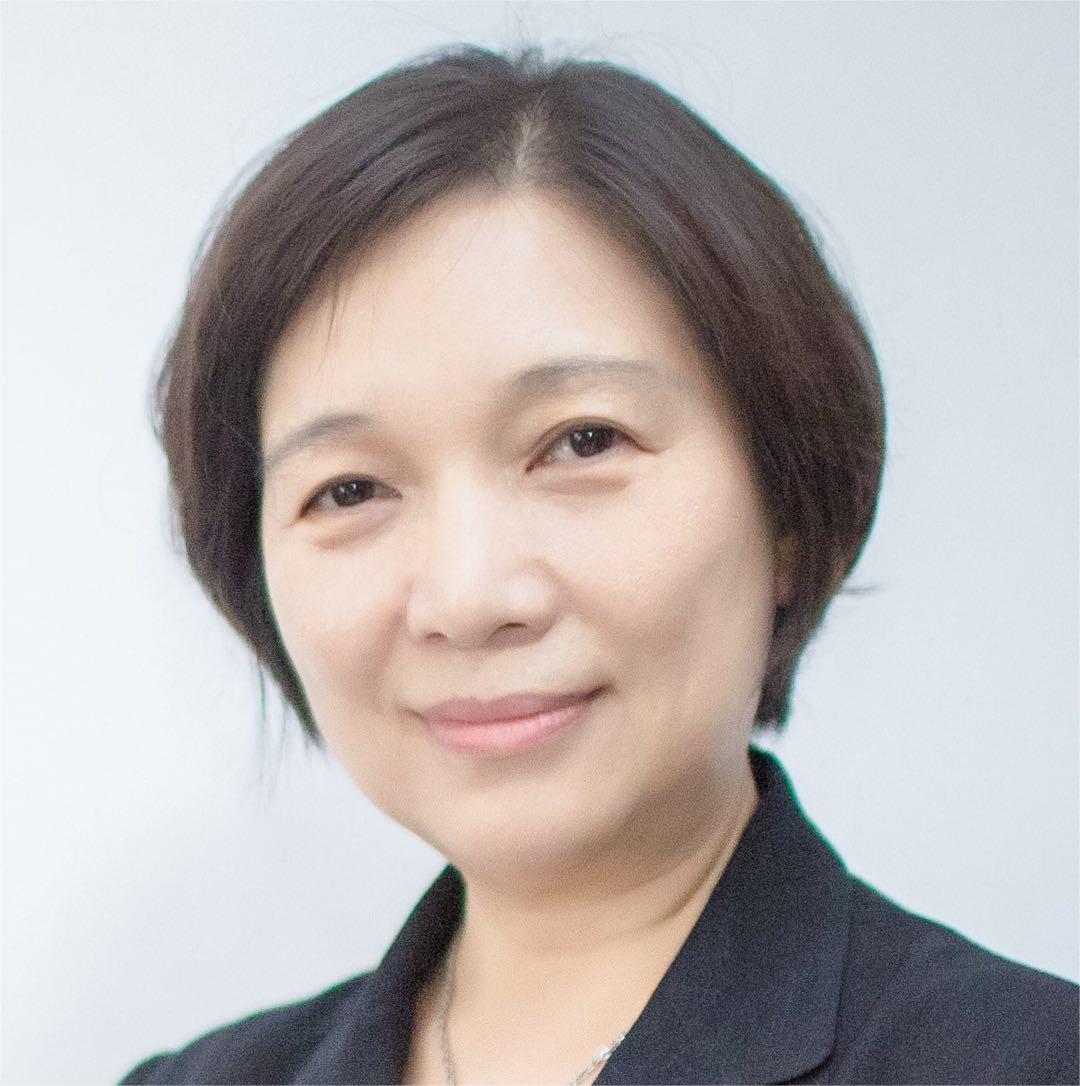}}},{}]
\noindent
\textbf{Jane Shen Shengmei} is the Chief Scientist of Pensees Singapore, and she is specialized 
in AI, Deep Learning, Face \& Image Recognition, 3D, Autonomous Driving, 
Image/video/audio Processing and Compression. Deep experience in leading research and 
technology teams in computer vision, AI and robotics domains, with highly cited publications 
in top journals/conferences and exceptional accomplishments in international competitions. 
Published and led research for over 150 papers and patents, with publications in venues 
including CVPR, NeurIPS, ICCV, ECCV, AAAI, CoRL, ICIP, ICPR, and Google Scholar 
profile of over 4800 citations, h-index of 38 and i10 index of 91. Directed team research and 
provided hands-on technical expertise for computer vision algorithm design that resulted in 
1st place results in PASCAL VOC 2010, 2011, 2012 PASCAL VOC, 1st place result in Visual 
Object Tracking in 2013, 1st place result in Microsoft 1M-Celebrity facial recognition 
competition 2017 (track 1 and 2), 1st place in anomaly detection track for CVPR 2018 AI 
City Challenge, 1st place for IROS 2018 mobile robotics challenge, 1st place for CVPR 2019 
lightweight facial recognition challenge across all 3 tracks. Research work directly translated 
into industry applications through Panasonic/Pensees products and solutions. Recognized
thought leader in the Asia-Pacific region and in Singapore for industry contributions in 
computer vision, AI and machine learning products. Awarded inaugural 100 Women in Tech 
Award 2020 by Singapore government, IT Awards Leader 2021 in Entrepreneurship by 
Singapore Computer Society.
\end{figwindownonum}

\end{document}